\documentclass[conference]{IEEEtran}
\IEEEoverridecommandlockouts
\usepackage{cite}
\usepackage{amsmath,amssymb,amsfonts}
\usepackage{algorithmic}
\usepackage{graphicx}
\usepackage{makecell, multirow}
\usepackage{hyperref}
\usepackage{textcomp}
\usepackage{xcolor}
\def\BibTeX{{\rm B\kern-.05em{\sc i\kern-.025em b}\kern-.08em
    T\kern-.1667em\lower.7ex\hbox{E}\kern-.125emX}}
\IEEEoverridecommandlockouts
\begin{document}

\title{Mamba-Enhanced Implicit Motion Learning for Audio-Driven Portrait Animation}

\author{
    \IEEEauthorblockN{
        Xuan Wei\textsuperscript{1, \dag}, 
        Jiahui Chen\textsuperscript{1, \dag}, 
        Kaiheng Li\textsuperscript{1}, 
        Mingyu Shao\textsuperscript{1}, 
        Qingqi Hong\textsuperscript{1, 2, 3, *}
    }
    \IEEEauthorblockA{
        \textsuperscript{1}\textit{Department of Digital Media Technology, School of Film, Xiamen University}, Xiamen, China \\
        \textsuperscript{2}\textit{National Institute for Data Science in Health and Medicine, Xiamen University}, Xiamen, China\\
        \textsuperscript{3}\textit{Institute of Artificial Intelligence, Xiamen University}, Xiamen, China \\
       Email: hongqq@xmu.edu.cn
    }
    \thanks{\textsuperscript{\dag} Equal contribution.}
    \thanks{\textsuperscript{*} Corresponding author. 
    This work was supported in part by the National Natural Science Foundation of China under Grant 62471418 and Fujian Provincial Natural Science Foundation of China under Grant 2024J01058, and the Giant Interactive Group Inc.}
}
\maketitle
\begin{abstract}
Audio-driven human motion video generation aims to synthesize realistic and temporally coherent human animations from a single static image, with applications in talking-head synthesis, co-speech gesture generation, and dynamic presentations. Moving beyond conventional keypoint-based methods that often struggle to capture subtle motion dynamics, We propose a novel implicit-motion framework for generating realistic and temporally coherent human motion videos from a single static image and audio. Our approach uses a two-stage pipeline that decouples motion prediction from rendering. The first stage integrates appearance priors and hierarchical depth cues into a region-aware attention mechanism to model latent motion features. The second stage employs a Mamba-enhanced diffusion model to directly predict these features from audio and the source image, enabling unsupervised learning of fine-grained motion patterns. This decoupled architecture enhances flexibility and efficiency. Trained on a new 380-hour high-quality dataset, our method outperforms prior work across multiple public benchmarks and our collected data in accuracy, naturalness, and temporal coherence, setting a new state-of-the-art. 
\end{abstract}


\section{Introduction}

\label{sec:intro}
In recent years, generative models like GANs and diffusion models have significantly advanced portrait animation, enabling the synthesis of dynamic videos from static images and audio. However, generating high-quality videos with sustained visual coherence remains challenging, as existing methods struggle with computational cost, stability, and generalization—especially under significant pose variations.

To address this, we propose a novel two-stage framework based on unsupervised implicit motion representation. For Stage 1, we first train a motion feature encoder-decoder network on video frame sequences to capture implicit motion deviations. These deviations are combined with the source image's appearance and depth features to produce coherent output. The core innovation is our Deviation Image Transformer (DIT), which fundamentally rethinks motion representation by generating spatial motion offsets through 3D convolution and adaptive normalization, effectively eliminating landmark artifacts. Subsequently, our Latent Motion Deviation Decoder (LMDD) achieves stratified motion control through depth-aware feature layering, coordinating deviation maps via softmax-based probability weighting to dynamically prioritize motion regions and resolve occlusion artifacts.
For Stage 2, our enhanced diffusion model synthesizes motion features during inference by integrating global/local audio features and motion priors from preceding frames, replacing the motion encoder to improve generation stability.

In summary, this work makes three key contributions:
\begin{itemize}
    \item We propose a novel implicit multimodal perception-based cross-domain generative framework for human portrait generation, overcoming the limitations of traditional explicit keypoint modeling and single-modality driven approaches.
    \item An unsupervised implicit motion representation learning framework is developed with the integration of a latent motion-region attention mechanism, which is further enhanced by a diffusion model using Mamba features to synthesize stable, long-duration human motion videos.
    \item We build a 380-hour high-quality audio-visual dataset for a two-stage training paradigm, where Stage 1 learns motion deviations from video frames and Stage 2 integrates Mamba-enhanced diffusion with audio-temporal conditioning for long-term coherence.
\end{itemize}

\section{Related Work}
\label{sec:related}  

\subsection{Co-Speech Gesture Generation}
Co-speech gesture generation aims to produce natural gestures that complement speech. Early end-to-end regression methods~\cite{Ginosar_2019} struggled with the complex speech-gesture relationship. Later, GANs and diffusion models~\cite{AhujaLIM20, Zhu_2023_CVPR} were adopted for higher fidelity. Recent advances focus on diffusion-based architectures~\cite{tevet2022human} proposed MDM, a diffusion transformer architecture that directly synthesizes gestures from text inputs using classifier-free guidance to enhance motion diversity. Concurrently, \cite{yan2024exploring} explores realistic human motion synthesis using kinematic descriptors within a diffusion framework, aiming to mitigate limb deformation artifacts common in generative models. For video generation, ANGIE~\cite{liu2022audio} employs MRAA features~\cite{Siarohin2021motion}, VQ-VAE~\cite{van2017neural} quantization, and a GPT-like network for predicting motion patterns, but its linear motion modeling is limited. To address this, DiffTED~\cite{Hogue_2024_CVPR} and S2G-MDDiffusion~\cite{he2024co} introduce a motion decoupling strategy, using the thin-plate spline (TPS)~\cite{Zhao_2022_tpsmm} to generate keypoints for rendering. Yet, this can lead to unnatural deformations and ghosting, due to limited influence of edge control points on the deformation field. TANGO~\cite{liu2024tango} presents CLIP-like embedding space between audio-motion modalities, which enhances the generalization potential of model. Consequently, we propose an implicit motion with latent attention deviation module aimed at enabling accurate representation for enhanced generation quality.

\subsection{Talking Head and Lip-syncing}
Talking head generation aims to synthesize a talking portrait video given the driving condition. Due to the complex motions the real videos contain, some talking head generation models~\cite{Chen_2019_CVPR} typically adopt facial landmarks as intermediate representations for video frames generation. Alternatively, some methods~\cite{corona2024vlogger} process head pose and lip synchronization in the latent space. But these methods face the problem of vagueness and inconsistency of the generated output videos. Meanwhile, contemporary methods~\cite{Du_2023_dae,Shen_2023_CVPR} integrate conditional control unit with UNet to achieve highly realistic facial renderings. Although these approaches achieve a fair degree of alignment between lip movements and audio, diffusion models still struggle to capture desirable motion and identity information, and generate a wide array of diverse outputs. Recent approach~\cite{guo2024liveportrait} explore implicit keypoint-based video-driven frameworks, using the scalable motion transformation. However, it struggles with cross-reenactment involving significant pose changes and shows jitter with pronounced shoulder movements in the input video.

\section{Method}
\label{sec:method}

\begin{figure*}
    \centering
    \includegraphics[width=0.7\linewidth]{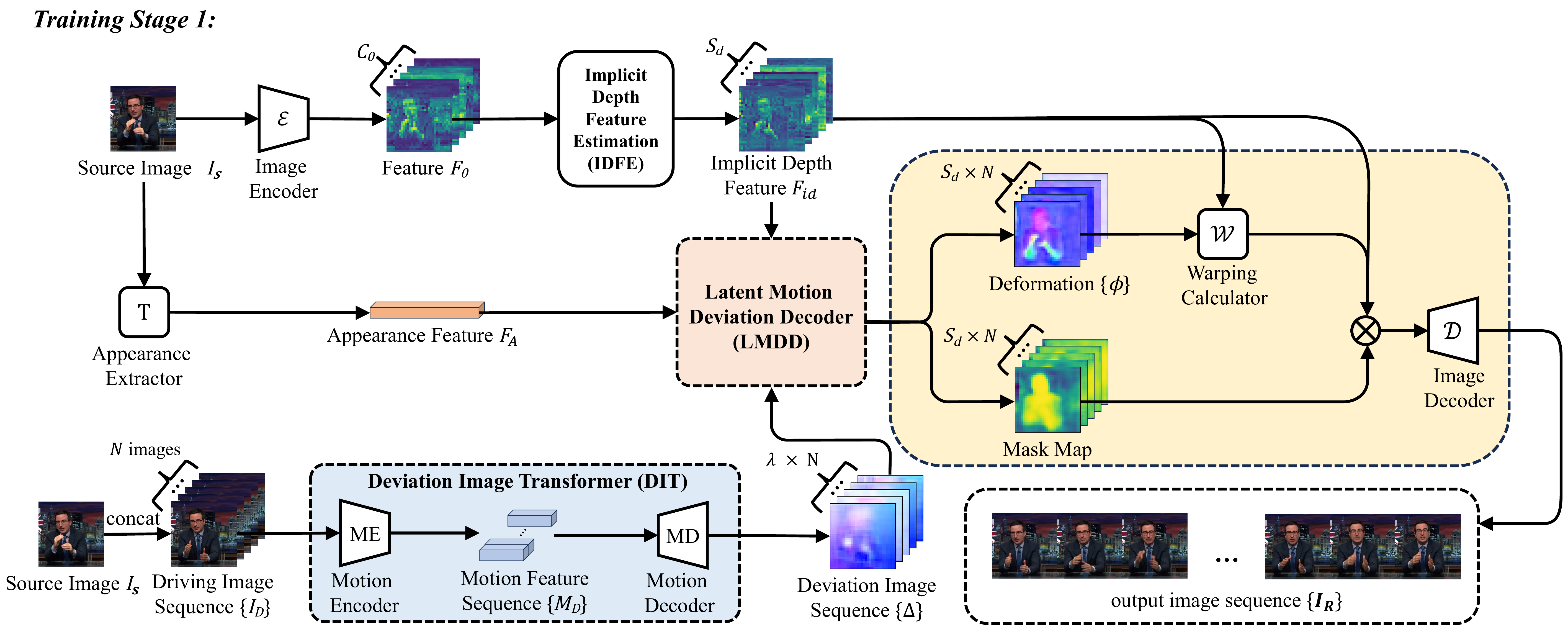}
    \caption{\textbf{Training stage 1: base model learning.} The image encoder $\mathcal{E}$, the deviation image transformer, the warping calculator ${w}$ and the image decoder $\mathcal{D}$ are learnable. In this stage, the model is trained from scratch. $C_S$ denotes the channels of feature, while $D_F$ represents the layer of depth.}
    \vspace{-0.3cm}
    \label{fig:stage1}
\end{figure*}

\begin{figure}
    \centering
    \includegraphics[width=0.9\linewidth]{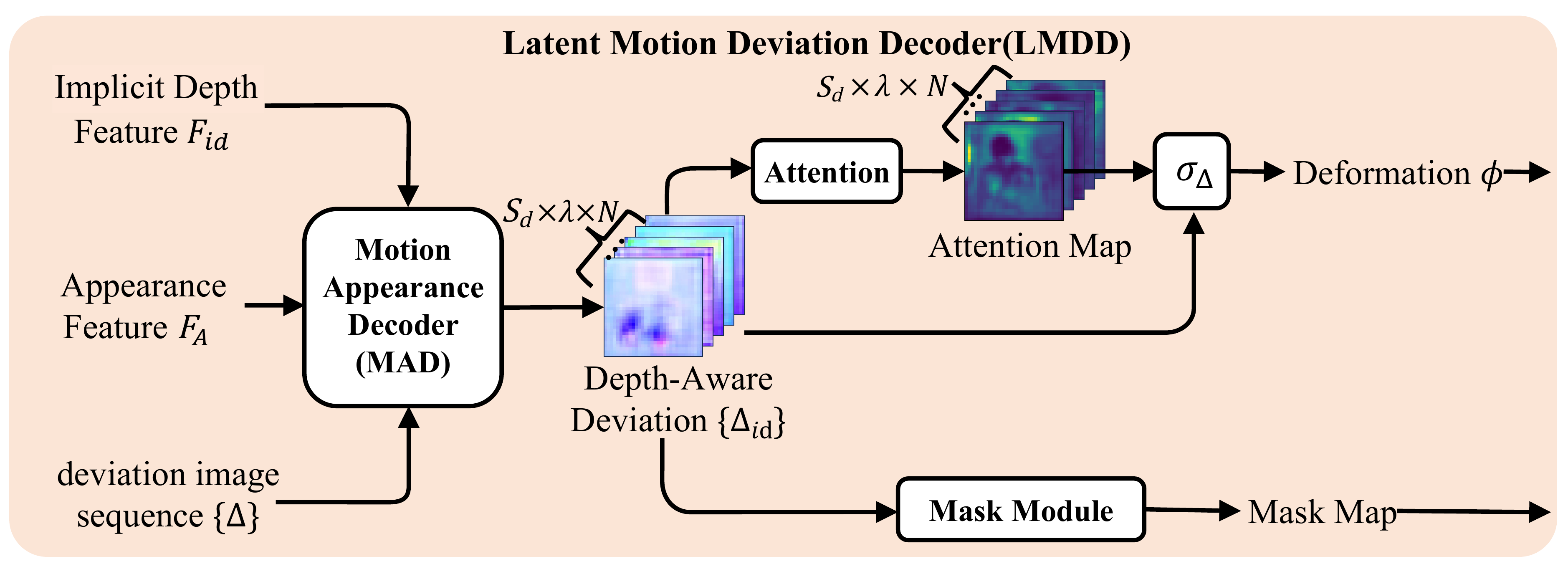}
    \caption{\textbf{The structure of Latent Motion Deviation Decoder.} The attention module is used to focus on the regions of interest for motion, while the mask module is employed to mask the range of motion.}
    \label{fig:LMDD}
     \vspace{-0.5cm}
\end{figure}

We propose a novel audio-driven human motion generation framework based on a full-scene deviation strategy. Unlike traditional methods that rely on explicit keypoints or 3D models, our approach learns an implicit motion representation to capture complex dynamics. The generation pipeline takes as input an audio signal $a$ and a source image $I_S$.

Our two-stage training framework (Fig. \ref{fig:stage1}) decouples motion prediction from image decoding for efficiency. Stage 1 trains a base model using a source image $I_S$ and driving sequence ${I_D}$. It includes a Deviation Image Transformer (DIT) to ensure temporal continuity and a Latent Motion Deviation Decoder (LMDD) for frame consistency. Stage 2 employs a Latent Motion Appearance Diffusion module to generate realistic motion features from audio $a$ and motion features ${MF_j}$. These features are processed by the pre-trained base model for final inference, reducing computational cost while improving generation quality.

\subsection{Training Stage 1: Base Model Learning}
We generate realistic human motion by mapping images to a latent space, rather than relying on traditional keypoint prediction. The core goal of this first stage is to learn a robust mapping from motion features to pixel space, establishing a foundation for generating coherent and identity-preserving video frames. While keypoint-based methods offer precise motion representation, they often struggle with constrained motion areas, coherence, and diversity. Our approach overcomes these limitations, enabling more natural and diverse motion generation.

\textbf{Implicit Depth Feature Estimation.} Human motion shows depth-dependent characteristics: proximal parts (e.g., fingers) exhibit larger displacements and higher velocities, while distal regions (e.g., torso) follow smoother paths. To handle this hierarchical scaling, we introduce a depth axis for motion stratification, using $S_d$ sectioning planes perpendicular to it.

The image encoder $\mathcal{E}$ processes the source image $I_S$ by first extracting an initial feature map $F_0 \in \mathbb{R}^{H \times W \times C_0}$, then convolves and expands it along the depth axis to produce $F_m \in \mathbb{R}^{H \times W \times C \times S_d}$, thereby creating $S_d$ discretely stratified depth layers. These depth-partitioned features undergo global pooling along the depth axis to produce compact spatial representations, which is then transformed through fully connected layers and normalized via Sigmoid activation to generate the final set of depth-wise weighting coefficients $\omega_m$.
\begin{equation}
    \omega_{m} = \sigma\left( \text{FC}\left( \frac{1}{S_d\times H \times W} \sum_{s=1}^{S_d} \sum_{h=1}^{H} \sum_{w=1}^{W} F_{m} \right) \right)
\end{equation}
These weights refine the features via a ResNet3D module, outputting the depth-aware representation $F_{id} = \omega_m\cdot F_m$.  

We propose an Appearance Extractor to address the issue of conventional displacement fields incorrectly transferring identity features (e.g., moles, wrinkles). It encodes biometric details from $I_S$ into a compact vector $F_A$, preserving identity while minimizing artifact interference for fusion.

\begin{figure*}
    \centering
    \includegraphics[width=0.7\linewidth]{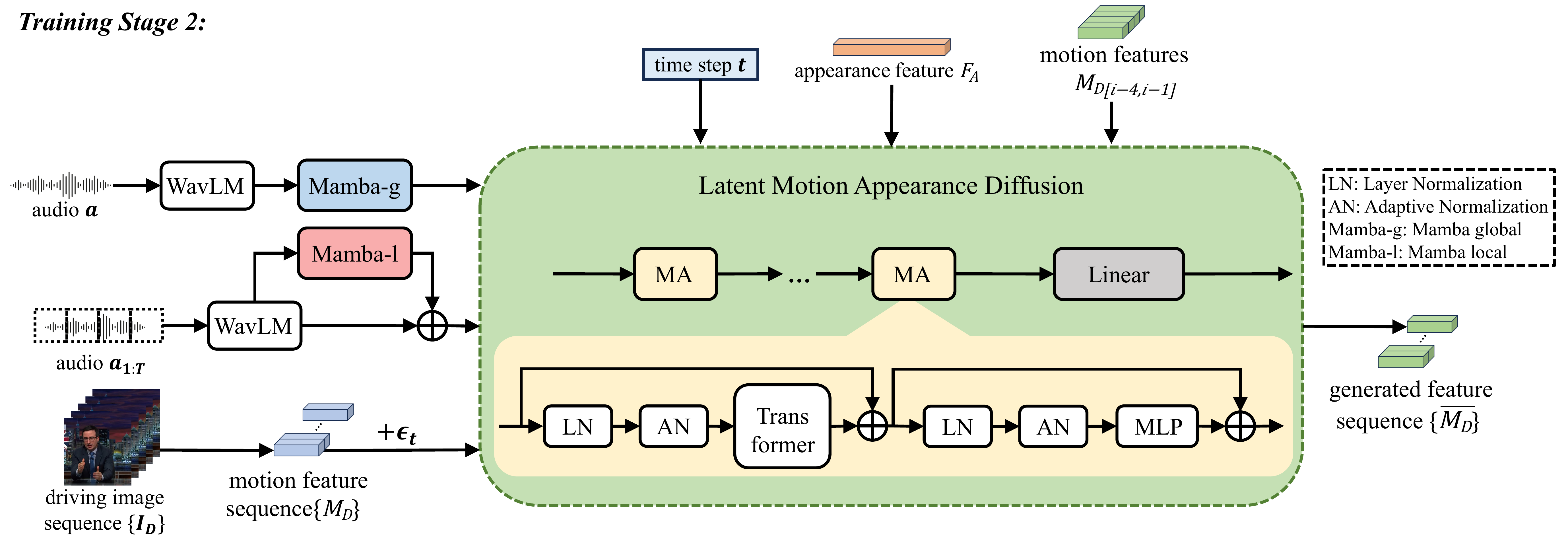}
    \caption{\textbf{Training stage 2: latent motion appearance diffusion training.} Global and local features are extracted from the audio by Mamba global extractor (Mamba-g) and Mamba local extractor (Mamba-l), and a Transformer-based Diffusion model is used to train the generation of motion features.}
    \label{fig:stage2}
    \vspace{-0.3cm}
\end{figure*}

\textbf{Deviation Image Transformer.} To address jitter artifacts inherent in traditional keypoint-based methods during large motions while enabling robust conversion from motion features to spatial deformation deviations, we propose the Deviation Image Transformer (DIT) which explicitly capture deformation details of driving images relative to the source image. The implementation begins by feeding source image $I_S$ and driving sequence $\{I_D\}$ into a Motion Encoder to output motion features $\{M_D^i\}^{N}_{i=0} \in \mathbb{R}^{768 \times (N+1)}$ (with $M_D^0$ corresponding to $I_S$). Following this, each $M_D^i$ is reshaped into tensor $V_{D}^i \in \mathbb{R}^{8\times 8\times 12}$ and normalized to standardized space representation $V_{nor}^i$. We then apply $n$ layers of 3D convolution to $V_{nor}^i$. Then scaling/shifting parameters $k_{sc}^i$, $k_{sh}^i$ decomposed from $k$ enable computation of the $i$-th deviation map: 
\begin{equation}
    \Delta_i=\frac{V_{nor}^i-V_{nor}^0}{\sqrt{\delta^2+\epsilon}}\cdot k_{sc}^i+k_{sh}^i,
\end{equation}
where $\delta$ is standard deviation and $\epsilon=10^{-5}$ prevents division by zero. To model independent motion in multiple regions, we generate $\lambda$ deviation maps per frame $\{\{\Delta_j\}_i\}$ ($i=1,\dots,N; j=1,\dots,\lambda$, where $\lambda$ denotes the number of motion regions of interest), denoted $\{\Delta\}$, providing pixel-level guidance for deformations through implicitly represented motion displacement fields.

\textbf{Latent Motion Deviation Decoder.} 
This module addresses motion-depth decoupling and attention-guided deformation by fusing depth information, motion deviations, and identity features into hierarchical deformation parameters, making the movement of the characters more natural. Motion deviations $\{\Delta\}$ decode depth features $F_{id}$ to produce depth-aware deviation sequences $\{\Delta_{id}\}$ where each element aligns with $S_d$ depth layers, forming $\mathbb{R}^{H\times W\times S_d}$ structures. The attention mechanism processes $\{\Delta_{id}\}$ to generate focus attention maps that highlight priority motion regions. By applying softmax $\sigma_\Delta$ across the $\lambda$ motion regions, it computes probability distributions for precise motion control, ultimately integrating these regions of interest to produce $N$ frames of deformation features $\{\phi\}\in\mathbb{R}^{H\times W\times S_d\times N}$. In parallel, the mask module generates per-layer motion region masks to provide spatial constraints.

\textbf{Image Decoder.} Before feeding the depth deviation features into the decoder for final prediction, we enhance the current features through a multi-stage process. Deformation information $\phi$ is first warped with depth features $F_{id}$ to produce $\phi'$, improving inter-frame deformation realism. Using a mask $\delta_\phi$, we then compute $\phi_m = \delta_\phi \phi' + (1-\delta_\phi)F_{id}$ to handle occlusions. Next, a nonlinear activation $\phi_{r} = \max(0,\phi_m) + \beta \cdot \min(0,\phi_m)$ modulates negative values, where $\beta$ controls articulation intensity. Finally, $\phi_r$ is fed into decoder $U$ to generate the output frame $I_R = U(\phi_r)$.

\textbf{Loss.} Following~\cite{Siarohin2021motion}, we use a pre-trained VGG-19 network to compute the multi-resolution perceptual reconstruction loss $\mathcal{L}_{per}$ between the generated image $I_R$ and ground truth $I_D$. For realism, a patch-based discriminator is trained with the adversarial loss $\mathcal{L}_{discr}$. Additionally, we propose a novel threshold mask loss: a binary mask $\mathrm{l_{mask}}$ is generated by thresholding ($\tau$) the mean absolute difference between ground truth features $v^{gt}$ and source deviation features $v^{dev}$, i.e., $\mathrm{l_{mask}} = \mathrm{mean}(|v^{gt} - v^{dev}|) > \tau$. This mask is then used to compute a weighted perceptual-geometric loss  $\mathcal{L}_{pg} = \sqrt{\mathrm{mean}((v^{gt} - v^{dev})^2)}\cdot \mathrm{l_{mask}}$, which ensures that only the differences exceeding the threshold contribute to the loss. 

Therefore, the total loss in stage 1 is as follows, where $\alpha_i$ is a hyperparameter:
\begin{equation}
\mathcal{L}_{total}^{Stage1} = \alpha_{1}\mathcal{L}_{per} + \alpha_{2}\mathcal{L}_{adv}^{G} + \alpha_{3}\mathcal{L}_{pg}
\end{equation}

\subsection{Training Stage 2: Motion Feature Generating}
Building upon the foundational image prediction framework, we achieve video generation by directly predicting motion feature sequences from audio inputs and source images, thereby decoupling the task and accelerating inference. To this end, we adapt the latent motion diffusion model from ~\cite{he2024co}. As shown in Fig.~\ref{fig:stage2}, this model is trained on motion feature sequences ${M_D}$—encoded from driving image sequences ${I_D}$—that are perturbed by noise $\epsilon_t$ at diffusion timestep $t$. We further incorporate multiple conditioning signals to regulate the synthesis.

Firstly, the appearance feature $F_A$ extracted from source image $I_S$ preserves identity consistency between generated frames and the original portrait. For audio conditioning, hierarchical embeddings are derived through dual-path processing: global features $F_{ag}$ obtained from raw speech $a$ via a Mamba global extractor maintain cross-segment stylistic consistency, while local features $F_{al}$ captured from audio slices $a_{1:T}$ using a Mamba local extractor model intra-segment dynamic variations. This fuzzy inference strategy enables implicit continuous feature learning without explicit classification, ensuring motion sequences closely approximate real-image-encoded dynamics for enhanced temporal coherence.  

Additionally, predicted motion features $\overline{M_D}_{[i-4, i-1]}$ from four preceding frames serve as weakly supervised conditions for the $i$-th target frame (initialized as zero vectors for $i<0$). This leverages inherent temporal continuity to stabilize generation by suppressing frame skipping and motion jitter while enhancing coherence during audio transitions.  

The diffusion model thus predicts denoised motion features $\{\overline{M_D}\}$ from noisy inputs $\{M_D\} + \epsilon_t$ under composite conditioning $c = (F_{ag}, F_{al}, \overline{M_D}_{[i-4,i-1]}, F_{A})$. Training incorporates a mean squared error (MSE) loss for motion prediction:  
\begin{equation}
\mathcal{L}_{MSE} = \| M_D - \overline{M_D} \|^2_2
\end{equation}
supplemented by implicit velocity and acceleration constraints to ensure physical plausibility:  
\begin{equation}
\mathcal{L}_{impvel} = \frac{1}{M-1}\sum_{m=1}^{M-1}\|f_{m+1}(M_{D}) - f_{m+1}(\overline{M_{D}})\|^{2}_{2}
\end{equation}

% \begin{equation}\label{equ}
% \begin{split}
% \mathcal{L}_{accel} = \frac{1}{M-2}\sum_{m=1}^{M-2} \left\| a_{m}^{gt} - a_{m}^{pred} \right\|_{2}^{2}
% \end{split}
% \end{equation}
\begin{equation}\label{equ}
\begin{split}
\mathcal{L}_{accel} = & \frac{1}{M-2}\sum_{m=1}^{M-2} \Bigg\| \big( f_{m+2}(M_{D}) - f_{m+1}(M_{D}) \big) \\
& - \big( f_{m+2}(\overline{M_{D}}) - f_{m+1}(\overline{M_{D}}) \big) \Bigg\|^{2}_{2},
\end{split}
\end{equation}
where $f_{k}(x) = x^{(k)} - x^{(k-1)}$ denotes frame-wise motion differences~\cite{li2022cvpr}, and $M$ is the number of predicted frames. Our diffusion model $\epsilon_{\theta}$ is trained to predict the noise $\epsilon$ that was added to the motion features$M_D$ at timestep $t$, following the $\epsilon$-prediction parameterization. The training objective is: The composite training objective is as follows:  
\begin{equation}
\begin{split}
\mathcal{L}_{simple} = & \mathbb{E}_{M_D, t, \epsilon \sim \mathcal{N}(0,I)} \left[ \| \epsilon - \epsilon_{\theta}( \sqrt{\bar{\alpha}_t} M_D \right. \\
                       & \left. + \sqrt{1-\bar{\alpha}_t} \epsilon, t, c ) \|^2 \right]
\end{split}
\end{equation}
This integrated loss function preserves biometric identity while ensuring natural articulation of hands, lips, and head throughout generated sequences.

\begin{table*}[ht!]
\caption{Quantitative results on talking head task. The table lists the metric results for the MEAD dataset and our self-collected dataset. Bold indicates the best and underline indicates the second.}
\centering
\small
\resizebox{\textwidth}{!} {
\begin{tabular}{c c c c c c c c c c c c c c c}
\multirowcell{2}[-4pt]{Method} & \multicolumn{7}{c}{MEAD} & \multicolumn{7}{c}{DiverseHeads} \\
\cline{2-15}
& Headpose$\downarrow$ & PSNR$\uparrow$ & SSIM$\uparrow$ & FID$\downarrow$ & FVD$\downarrow$ & LSE-D$\downarrow$ & LSE-C$\uparrow$ & Headpose$\downarrow$ & PSNR$\uparrow$ & SSIM$\uparrow$ & FID$\downarrow$ & FVD$\downarrow$ & LSE-D$\downarrow$ & LSE-C$\uparrow$ \\
\hline
AniPortrait~\cite{wei2024aniportrait} & 1.476 & 27.446 & 0.880 & 64.755 & 218.258 & 10.430 & 1.408 & 1.920 & 23.245 & 0.804 & 42.941 & 225.998 & 10.040 & 1.605 \\
DaGAN~\cite{hong2022dagan} & 8.008 & 31.811 & 0.934 & 34.834 & 129.180 & 10.775 & 1.214 & 5.159 & 26.119 & 0.863 & 31.567 & 115.952 & 10.177 & 1.579 \\
FOMM~\cite{Siarohin_2019_FOMM} & 7.977 & 29.236 & 0.899 & 37.022 & 143.790 & 10.437 & 1.447 & 5.134 & 22.990 & 0.803 & 33.465 & 150.046 & 10.177 & 1.608 \\
LivePortrait~\cite{guo2024liveportrait} & \textbf{1.348} & 32.518 & \underline{0.941} & 34.125 & \underline{102.980} & 10.463 & 1.048 & \textbf{1.339} & 26.216 & 0.872 & \underline{25.401} & \underline{108.540} & \underline{9.814} & \underline{2.098} \\
MCNet~\cite{Hong_2023_ICCV} & 7.397 & \underline{32.804} & 0.940 & \underline{33.141} & 135.458 & \underline{10.038} & \underline{1.812} & 4.207 & \underline{27.759} & \underline{0.885} & 27.730 & 113.583 & 10.075 & 1.739 \\
TPSMM~\cite{Zhao_2022_tpsmm} & 7.874 & 31.924 & 0.940 & 34.619 & 136.075 & 10.486 & 1.370 & 5.148 & 26.997 & 0.879 & 27.389 & 121.976 & 10.096 & 1.717 \\
Ours & \underline{1.465} & \textbf{34.661} & \textbf{0.953} & \textbf{30.501} & \textbf{92.179} & \textbf{9.025} & \textbf{2.672} & \underline{1.826} & \textbf{29.213} & \textbf{0.908} & \textbf{23.718} & \textbf{77.267} & \textbf{9.248} & \textbf{2.777} \\
\hline
\end{tabular}
}
%\caption{Quantitative results on talking head task. The table lists the metric results for the MEAD dataset and our self-collected dataset. Bold indicates the best and underline indicates the second.}
\label{tab:talking_head}
\vspace{-0.3cm}
\end{table*}

\begin{table*}
\caption{Quantitative results on co-speech gesture task. Bold indicates the best and underline indicates the second.}
\centering
\small
\resizebox{0.8\textwidth}{!} {
\begin{tabular}{lcccccccccc}
    Method & Headpose$\downarrow$ & PSNR$\uparrow$ & SSIM$\uparrow$ & LPIPS$\downarrow$ & FID$\downarrow$ & FVD$\downarrow$ & DIV$\uparrow$ & TGD$\downarrow$ & LSE-D$\downarrow$ & LSE-C$\uparrow$ \\
    \hline
    S2G-MDDiffusion~\cite{he2024co} & \underline{1.333} & \underline{28.500} & \underline{0.894} & \underline{0.040} & \underline{25.325} & 109.424 & 6.280 & 6.610 & 11.010 & 0.971 \\
    TANGO~\cite{liu2024tango} & - & - & - & - & 42.863 & 168.313 & 5.217 & \textbf{1.883} & \underline{10.365} & \textbf{1.736} \\
    Ours & \textbf{1.115} & \textbf{28.835} & \textbf{0.905} & \textbf{0.027} & \textbf{19.246} & \textbf{54.456} & \textbf{7.159} & \underline{5.027} & \textbf{10.257} & \underline{1.531} \\
    \hline
    w/o Appearance & 1.694 & 26.299 & 0.869 & 0.043 & 28.194 & \underline{90.955} & \underline{6.511} & 6.887 & 11.260 & 0.721 \\
    w/o Trans. in DIT & 1.576 & 24.790 & 0.854 & 0.057 & 32.234 & 153.641 & 5.800 & 7.213 & 11.692 & 0.309 \\
    w/o attention map & 1.500 & 25.801 & 0.863 & 0.047 & 29.148 & 133.441 & 6.405 & 6.597 & 11.780 & 0.296 \\
    w/o mamba-g & - & - & - & - & 31.136 & 70.773 & 6.139 & 6.697 & 11.038 & 1.230 \\
    w/o mamba-l & - & - & - & - & 30.587 & 71.032 & 4.706 & 5.593 & 11.000 & 1.090 \\
    \hline
    \label{tab:co-speech}
    \vspace{-0.5cm}
    \end{tabular}
    }
\end{table*}

\section{Experiments}
\label{sec:exp}

\begin{table}
    \caption{The inference time for generating a one-second, 25-FPS video in img2img and audio2video scenarios.}
    \centering
    \small
    \resizebox{0.7\linewidth}{!} {
    %\small
    \begin{tabular}{l c c}
    Method & img2img (s) & audio2video (s) \\
    \hline
    X-Portrait~\cite{xie2024x} & 36.973 & - \\
    Hallo2~\cite{cui2024hallo2} & - & 22.088 \\
    FOMM~\cite{Siarohin_2019_FOMM}  & \textbf{0.828} & - \\
    LivePortrait~\cite{guo2024liveportrait} & \underline{1.127} & - \\
    AniPortrait~\cite{wei2024aniportrait} & 11.498 & \underline{11.816} \\
    DaGAN~\cite{hong2022dagan} & 2.126 & - \\
    MCNet~\cite{Hong_2023_ICCV} & 2.663 & - \\
    TPSMM~\cite{Zhao_2022_tpsmm} & 2.752 & - \\
    Ours & 1.322 & \textbf{1.515} \\
    \hline
    \end{tabular}
    }
    \label{tab:performance_comparison}
    \vspace{-0.5cm}
\end{table}

\subsection{Implementation Details}
\quad \textbf{Datasets.} Experiments for the talking-head task are conducted on multiple datasets: CREMA-D, RAVDESS, HDTF, MEAD, CMLR, and our self-collected 380-hour "DiverseHeads" dataset (covering diverse speaking scenarios, 380 hours). For co-speech gesture generation, we use the PATS dataset (~84k clips, 250 hours). We follow S2G-MDDiffusion's standard preprocessing for four individuals (Chemistry, Oliver, Jon, Seth) in PATS.

\textbf{Evaluation metrics.} 
We comprehensively evaluate human motion video generation using both image-based and motion-specific metrics. Image quality is assessed via FID, while video quality and temporal coherence are measured using FVD and Temporal Gesture Distance (TGD). Motion naturalness is evaluated with Headpose error, and gesture diversity via DIV. Audio-visual synchronization is quantified using LSE-C and LSE-D, where a lower LSE-D indicates higher consistency.

\subsection{Comparisons with Other Methods}
\begin{figure}
    \centering
    \includegraphics[width=0.9\linewidth]{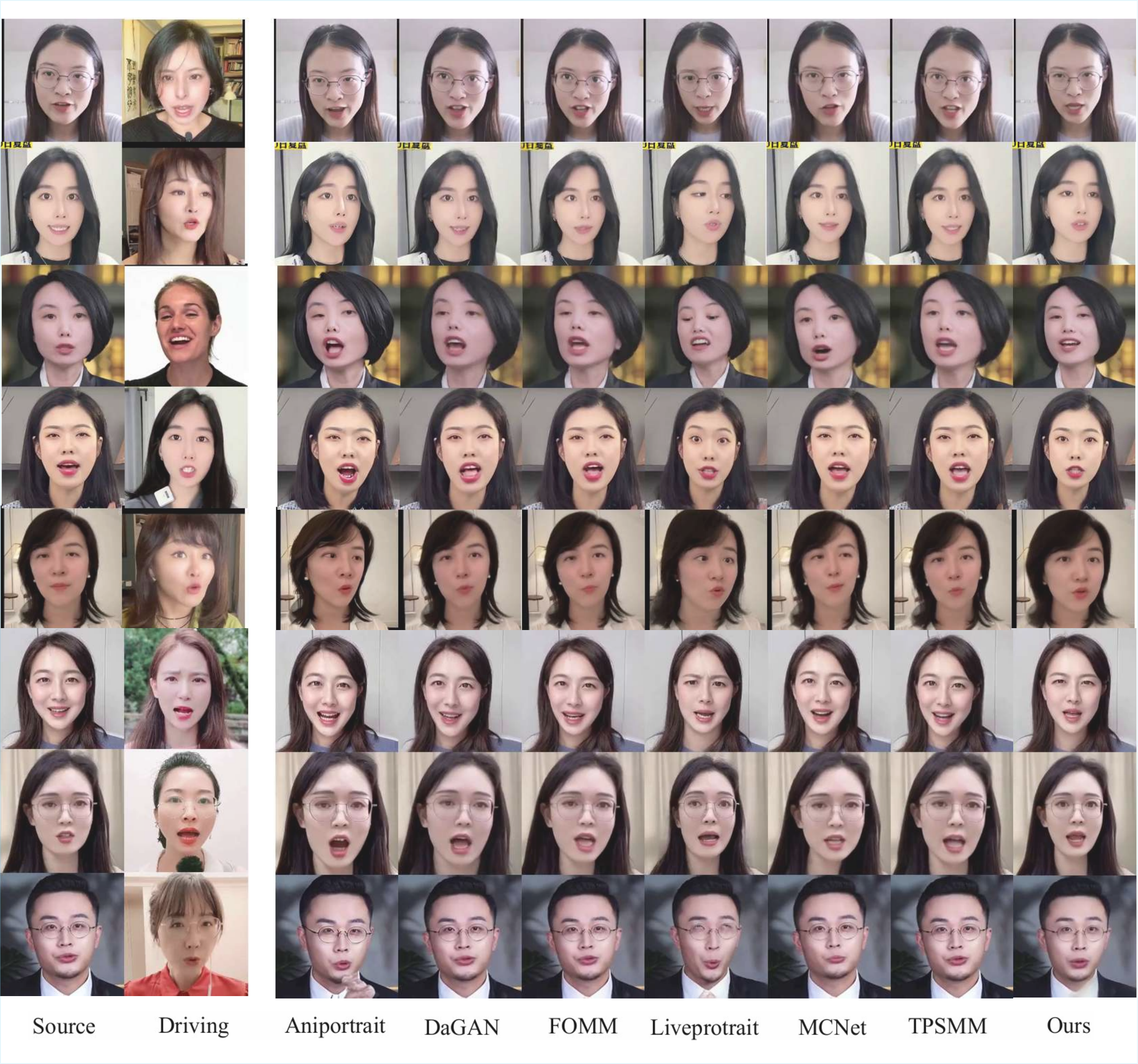}
    \caption{\textbf{Qualitative comparisons of talking head task.} We used different datasets for comparison. Our model preserves lip movements and eye gaze, handles large poses more stably, and better maintains the identity of the source portrait.}
    \label{fig:comparison}
    \vspace{-0.6cm}
\end{figure}
For the talking-head task, we compare our method with state-of-the-art approaches including AniPortrait~\cite{wei2024aniportrait}, DaGAN~\cite{hong2022dagan}, FOMM~\cite{Siarohin_2019_FOMM}, LivePortrait~\cite{guo2024liveportrait}, MCNet~\cite{Hong_2023_ICCV}, TPSMM~\cite{Zhao_2022_tpsmm}, and X-Portrait~\cite{xie2024x}. Using a driving video and a source portrait, all models generate videos at 256×256 resolution and 25 fps. As shown in Fig.~\ref{fig:comparison}, our method produces more consistent and realistic results compared with other methods. It exhibits higher similarity to the driving image, particularly in terms of head position and rotation angle, as well as the extent of lip movement.

On both MEAD and our DiverseHeads dataset (Table \ref{tab:talking_head}), our method achieves the best performance in most metrics. It is only slightly outperformed by LivePortrait on Headpose, which explicitly optimizes for this metric, demonstrating our approach's effectiveness without task-specific supervision.

\begin{figure}
    \centering
    \includegraphics[width=0.8\linewidth]{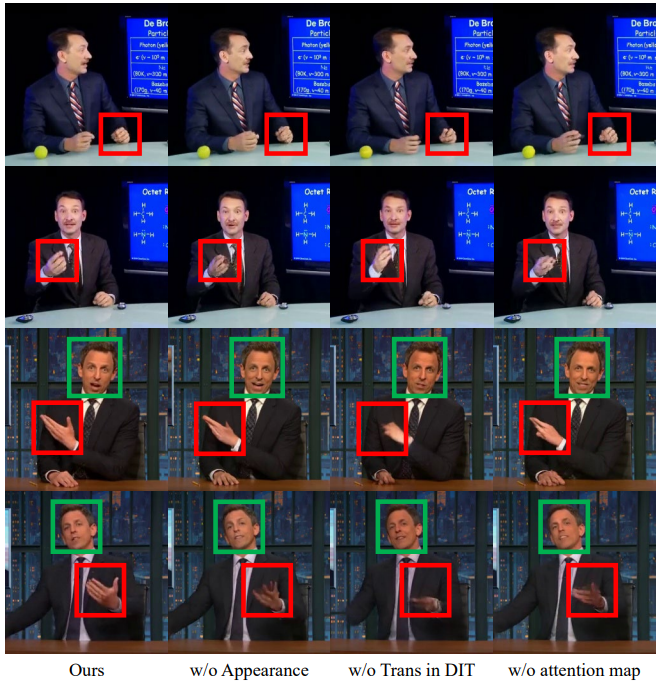}
    \caption{\textbf{Visualization results of the ablation study.} Red box and green box indicate the gesture and facial results generated by different models respectively  within the same frame.}
    \label{fig:ablation}
    \vspace{-0.6cm}
\end{figure}

For co-speech gesture generation on the PATS test set (256×256, 25fps), our method outperforms S2G-MDDiffusion across all metrics and performs comparably to TANGO on TGD and LSE-C (Table~\ref{tab:co-speech}). A fair comparison on metrics like Headpose or PSNR is not applicable to TANGO, as its pipeline relies on driving video frames and Wav2Lip for synthesis; this use of interpolated frames also limits its motion diversity. Our method produces clearer gestures and richer expressions, even with cross-identity inputs, highlighting its strong generalization. This capability stems from our framework's disentangled motion–appearance representation, where the LMDD models motion dynamics separately from identity features. Our method also demonstrates competitive efficiency, particularly in audio-driven generation (Table~\ref{tab:performance_comparison}).

\begin{table}[t]
    \centering
    \small
    \caption{The ablation studey on mamba extractors. We compare the full model (ours) with variants excluding the global (w/o mamba-g) or local extractors (w/o mamba-l) extractors.}
     \resizebox{0.9\linewidth}{!} {
    \label{tab:ablation}
    \begin{tabular}{lcccccc}
    \hline
    \textbf{Method} & \textbf{FID} $\downarrow$ & \textbf{FVD} $\downarrow$ & \textbf{DIV} $\uparrow$ & \textbf{TGD} $\downarrow$ & \textbf{LSE-D} $\downarrow$ & \textbf{LSE-C} $\uparrow$ \\
    \hline
    w/o mamba-g  & 31.136 & 70.773 & 6.139 & 6.697 & 11.038 & 1.230 \\
    w/o mamba-l  & 30.587 & 71.032 & 4.706 & 5.593 & 11.000 & 1.090 \\
    \textbf{Ours} & \textbf{19.246} & \textbf{54.456} & \textbf{7.159} & \textbf{5.027} & \textbf{10.257} & \textbf{1.531} \\
    \hline
    \end{tabular}
    \vspace{-0.6cm}
    }
\end{table}
\subsection{Ablation Study and Analysis}

The ablation studies (Fig.~\ref{fig:ablation}, Table~\ref{tab:co-speech}) validate the necessity of each component. Removing appearance features causes stiff motions, loss of hand details, and degrades Headpose accuracy. Omitting the Transformer in the DIT module leads to loss of motion details (Fig.~\ref{fig:ablation}, red box) and lowers PSNR, SSIM, and FVD scores. Without the attention map, lip and gesture details deteriorate, harming the LSE metric. Thus, both appearance features and the DIT-encoded motion features are indispensable for high-quality, consistent generation.

Similarly, ablating audio feature extraction in Mamba's second stage (Table~\ref{tab:co-speech}) shows that enhanced audio features, particularly global ones, substantially improve video quality and gesture metrics, confirming our design's effectiveness.

\textbf{Inference Efficiency.} 
We also evaluate inference efficiency across model configurations (Table~\ref{tab:ablation}). Compared to ablated variants without the global (\textit{w/o mamba-g}) or local (\textit{w/o mamba-l}) Mamba branch, our full model achieves the best perceptual quality (lowest FID/FVD, highest DIV/LSE-C) while maintaining competitive speed in generating 1-second, 25-FPS videos (both img2img and audio2video). This demonstrates that integrating both Mamba modules provides an effective quality-efficiency trade-off.

\section{Conclusion}
This paper presents a novel method for generating realistic and temporally consistent human motion videos from static portraits. Our model employs an implicit motion representation combined with a latent appearance-motion bias attention mechanism. Extensive experiments confirm its high-quality output and strong generalization.
Future directions aim to enhance realism: first, by decoupling facial animation control using physiological signals and expert-network submodules; second, by integrating neural radiance fields for dynamic, real-time rendering; and third, by developing a multimodal affective cognition model to drive naturalistic emotional expression. These developments will enable cinematic-quality digital humans for immersive communication.

\bibliographystyle{IEEEbib}
\bibliography{icme2026references}

\end{document}